# Fast Geometric Fit Algorithm for Sphere Using Exact Solution

Sumith YD
*Syracuse University, Mechanical and Aerospace engineering department*
*Syracuse, NY, USA 13244*
*syesudas@syr.edu*

Sphere fitting is a common problem in almost all science and engineering disciplines. Most of methods available are iterative in behavior. This involves fitting of the parameters in a least square sense or in a geometric sense. Here we extend the methods of Thomas Chan and Landau who fitted the 2D data using circle. This work closely resemble their work in redefining the error estimate and solving the sphere fitting problem exactly. The solutions for center and radius of the sphere can be found exactly and the equations can be hard coded for high performance. We have also shown some comparison with other popular methods and how this method behaves.

## I. INTRODUCTION

The fitting 3D data into a sphere is a desired problem in many disciplines including computer vision, molecular simulation, pattern recognition etc. There are mainly two methods to fit data into a given equation. The first one is fitting of characteristics equation of sphere in a least square fashion. The second is fitting the equation of sphere using geometric sense. The second method is preferred by many due to its accuracy and physical sense. There exists a wide range of literature for circle fitting for 2D data. Here we will limit ourselves to sphere fitting and 3D fitting related work. In 2000 Ahn et.al [1] has proposed a method to solve geometric fitting of sphere using Jacobian and matrix method. Their work was shown superior to that of Spath[2] and Gander[3]. In 2000 David Eberly [4] has come up with an iterative and efficient way to fit sphere onto 3D data.

There are a very few non iterative methods known for sphere fitting. In 1989 Thomas and Chan[5] has proposed a modified version of Landau[6] which uses exact solution for finding the fitting parameters. In 1990 Chaudhari[7] has proposed a variant of Thomas and Chan's [5] method, but the dependency on object area will make it unsuitable for partial sphere data.

In this work we will extend the methods of Thomas and Chan[5] to sphere fitting of 3D data. For comparison of the accuracy and speed of this method we will select the most widely accepted and used methods of Eberly[4] and Ahn[1] instead of comparing with the vast overwhelming literature.

## II. DERIVATION OF EXACT SOLUTION

Following Thomas and Chan[5], the estimation error can be considered as a difference between the sphere areas. For a given set of data points $(x_i, y_i, z_i)$ where $i = 1$ to $N$ the error can be defined as,

$$e = \sum_{i=1}^{N}\left[4\pi R^2 - 4\pi[(x_i - x_0)^2 + (y_i - y_0)^2 + (z_i - z_0)^2]\right]^2 \quad (1)$$

Redefining the error after removing the constant terms,

$$J = \sum_{i=1}^{N}[R^2 - (x_i - x_0)^2 - (y_i - y_0)^2 - (z_i - z_0)^2]^2 \quad (2)$$

For convenience let us define,

$$\Lambda = R^2 - (x_i - x_0)^2 - (y_i - y_0)^2 - (z_i - z_0)^2 \quad (3)$$

and $\Pi^2 = (x_i - x_0)^2 + (y_i - y_0)^2 + (z_i - z_0)^2 \quad (4)$

Minimizing the error $J(R, x_0, y_0, z_0)$ w.r.t the unknown parameters, $R, x_0, y_0$ and $z_0$ will yield

$$\partial J/\partial R = 2\sum R\Lambda = 0 \quad (5)$$

i.e., $\sum R^2 = \sum \Pi^2 \quad (6)$

$$\partial J/\partial x_0 = 2\sum \Lambda . -2(x_i - x_0). -1 = 0 \quad (7)$$

$$\sum \Lambda x_i = \sum \Lambda x_0 \quad (8)$$

In fact the RHS of the above equation is zero. Therefore

$$\sum \Lambda x_i = 0 \quad (9)$$

$$\sum (R^2 - \Pi^2)x_i = 0 \quad (10)$$

$$\sum R^2 x_i - \sum \Pi^2 x_i = 0 \quad (11)$$

Similarly minimizing for $y_0$ and $z_0$ will give the following equations.

$$\sum R^2 y_i - \sum \Pi^2 y_i = 0 \quad (12)$$

$$\sum R^2 z_i - \sum \Pi^2 z_i = 0 \quad (13)$$

Recall from equation 6
$$NR^2 = \sum \Pi^2 \quad (14)$$

Now multiply equation 14 with $\Sigma_x$, $\Sigma_y$ and $\Sigma_z$

$$NR^2\Sigma_x - \sum \Pi^2 \Sigma_x = 0 \tag{15}$$

$$NR^2\Sigma_y - \sum \Pi^2 \Sigma_y = 0 \tag{16}$$

$$NR^2\Sigma_z - \sum \Pi^2 \Sigma_z = 0 \tag{17}$$

Subtract equations 11-13 from 15-17 after multiplying with $N$.

$$N\sum \Pi^2 x_i - \sum \Pi^2 \Sigma_x = 0 \tag{18}$$

$$N\sum \Pi^2 y_i - \sum \Pi^2 \Sigma_y = 0 \tag{19}$$

$$N\sum \Pi^2 z_i - \sum \Pi^2 \Sigma_z = 0 \tag{20}$$

The next step is to simply the equations. The set of nonlinear equations can be solved linearly if we rearrange and simplify the terms.

$$\Pi^2 = (x_i^2 + y_i^2 + z_i^2) - 2x_i x_0 - 2y_i y_0 - 2z_i z_0 + (x_0^2 + y_0^2 + z_0^2) \tag{21}$$

$$\sum \Pi^2 = \Sigma_{x^2} + \Sigma_{y^2} + \Sigma_{z^2} - 2x_0\Sigma_x - 2y_0\Sigma_y - 2z_0\Sigma_z + N(x_0^2 + y_0^2 + z_0^2) \tag{22}$$

$$\sum \Pi^2 x_i = \Sigma_{x^3} + \Sigma_{xy^2} + \Sigma_{xz^2} - 2x_0\Sigma_{x^2} - 2y_0\Sigma_{xy} - 2z_0\Sigma_{xz} + (x_0^2 + y_0^2 + z_0^2)\Sigma_x \tag{23}$$

$$\sum \Pi^2 y_i = \Sigma_{x^2 y} + \Sigma_{y^3} + \Sigma_{yz^2} - 2x_0\Sigma_{xy} - 2y_0\Sigma_{y^2} - 2z_0\Sigma_{yz} + (x_0^2 + y_0^2 + z_0^2)\Sigma_y \tag{24}$$

$$\sum \Pi^2 z_i = \Sigma_{x^2 z} + \Sigma_{y^2 z} + \Sigma_{z^3} - 2x_0\Sigma_{xz} - 2y_0\Sigma_{yz} - 2z_0\Sigma_{z^2} + (x_0^2 + y_0^2 + z_0^2)\Sigma_z \tag{25}$$

With these values, equations 18-20 will become

$$N\sum \Pi^2 x_i - \sum \Pi^2 \Sigma_x = (N\Sigma_{x^3} + N\Sigma_{xy^2} + N\Sigma_{xz^2} - \Sigma_{x^2}\Sigma_x - \Sigma_{y^2}\Sigma_x - \Sigma_{z^2}\Sigma_x) + x_0(-2N\Sigma_{x^2} + 2\Sigma_x^2) + y_0(-2N\Sigma_{xy} + 2\Sigma_x\Sigma_y) + z_0(-2N\Sigma_{xz} + 2\Sigma_x\Sigma_z) = 0 \tag{26}$$

$$N\sum \Pi^2 y_i - \sum \Pi^2 \Sigma_y = (N\Sigma_{x^2 y} + N\Sigma_{y^3} + N\Sigma_{yz^2} - \Sigma_{x^2}\Sigma_y - \Sigma_{y^2}\Sigma_y - \Sigma_{z^2}\Sigma_y) + x_0(-2N\Sigma_{xy} + 2\Sigma_x\Sigma_y) + y_0(-2N\Sigma_{y^2} + 2\Sigma_y^2) + z_0(-2N\Sigma_{yz} + 2\Sigma_y\Sigma_z) = 0 \tag{27}$$

$$N\sum \Pi^2 z_i - \sum \Pi^2 \Sigma_z = (N\Sigma_{x^2 z} + N\Sigma_{y^2 z} + N\Sigma_{z^3} - \Sigma_{x^2}\Sigma_z - \Sigma_{y^2}\Sigma_z - \Sigma_{z^2}\Sigma_z) + x_0(-2N\Sigma_{xz} + 2\Sigma_x\Sigma_z) + y_0(-2N\Sigma_{yz} + 2\Sigma_y\Sigma_z) + z_0(-2N\Sigma_{z^2} + 2\Sigma_z^2) = 0 \tag{28}$$

Equations 26-28 are of the form

$$-d + ax_0 + by_0 + cz_0 = 0 \tag{29}$$

$$-h + ex_0 + fy_0 + gz_0 = 0 \tag{30}$$

$$-m + jx_0 + ky_0 + lz_0 = 0 \tag{31}$$

We can solve these equations using Cramer's rule[8] to obtain center of the circle.

$$x_0 = \frac{d(fl - gk) - h(bl - ck) + m(bg - cf)}{a(fl - gk) - e(bl - ck) + i(bg - cf)} \tag{32}$$

$$y_0 = \frac{a(hl - mg) - e(dl - mc) + j(dg - hc)}{a(fl - gk) - e(bl - ck) + i(bg - cf)} \tag{33}$$

$$z_0 = \frac{a(fm - hk) - e(bm - dk) + j(bh - df)}{a(fl - gk) - e(bl - ck) + i(bg - cf)} \tag{34}$$

Substituting in equation 6 we can get the radius

$$R^2 = \frac{1}{N}\left[\Sigma_{x^2} + \Sigma_{y^2} + \Sigma_{z^2} - 2x_0\Sigma_x - 2y_0\Sigma_y - 2z_0\Sigma_z\right] + (x_0^2 + y_0^2 + z_0^2) \tag{35}$$

It is notable that these equations are straight forward to implement in computer programs and the hardcoded version will give a good computational efficiency.

## III. RESULTS AND COMPARISON

In this section we will discuss and compare the accuracy and execution speed of the present method with two well-known iterative methods Eberly [4] and Ahn [1]. We will test these methods with a variety of data as shown in Table 1. Case 1 & 2 are spheres, Case 2 is hemisphere and Case 4 is a small portion of sphere. These are obtained by changing the parameter u in the parametric form of sphere equation as given below.

$$x = x_0 + \sqrt{r^2 - u^2} \cos\theta \tag{36}$$

$$y = y_0 + \sqrt{r^2 - u^2} \sin\theta \tag{37}$$

$$z = z_0 + u \tag{38}$$

Table I: Fitting parameters and case studies

|        | $x_0$  | $y_0$  | $z_0$   | R       | $\varepsilon$ | u        |
|--------|--------|--------|---------|---------|------|----------|
| Case 1 | 1      | 2      | 3       | 7.2     | 0.1  | -1 to 1  |
| Case 2 | 2.3423 | 0.8764 | 45.8785 | 9.02321 | 0.2  | -1 to 1  |
| Case 3 | 2.3423 | 0.8764 | 45.8785 | 9.02321 | 0.12 | -1 to 0  |
| Case 4 | 2.3423 | 0.8764 | 45.8785 | 9.02321 | 0.12 | -1 to -0.5 |

For each case, 100 data points will be generated using the random number generator. A random noise parameter $\varepsilon$ will be added to all data points to make the data less predictable. The parameter estimation will be done for this data using all methods. This procedure is repeated for another 1500 data set (100 x 1500 data points) to make sure we have considered enough random data. Preliminary tests shown that the method by Ahn [1] is computationally inefficient. Hence we will discuss only the results from Eberly's method[4]. The results of parameter estimation for the present method is shown in Table II.

Table II: Fitted values from present study

|  | $x_0$ | $y_0$ | $z_0$ | R |
|---|---|---|---|---|
| Case 1 | 0.9999 | 1.9993 | 2.9995 | 7.2006 |
| Case 2 | 2.3422 | 0.8763 | 45.8781 | 9.0254 |
| Case 3 | 2.3400 | 0.8735 | 45.8705 | 9.0214 |
| Case 4 | 2.3378 | 0.8720 | 45.8237 | 8.9837 |

For Eberly's method we have used an error tolerance of 0.0001 and maximum ot 25 steps for convergence. The results of estimation using this method for all the cases are shown in Table III.

Table III: Fitted values using Eberly's method [4]

|  | $x_0$ | $y_0$ | $z_0$ | R |
|---|---|---|---|---|
| Case 1 | 1.0002 | 1.9997 | 3.0002 | 7.2000 |
| Case 2 | 2.3424 | 0.8763 | 45.8786 | 9.0240 |
| Case 3 | 2.3431 | 0.8752 | 45.5744 | 8.8622 |
| Case 4 | 2.3381 | 0.8737 | 43.9396 | 7.6262 |

The fitted values shown good accuracy for sphere and least for the portion of the sphere. This can be estimated using the RMS error value of fit.

$$RMS_{MAX} = MAX\left[\sum \frac{(x_i-x_0)^2}{N}, \sum \frac{(y_i-y_0)^2}{N}, \sum \frac{(z_i-z_0)^2}{N}, \sum \frac{(R_i-R)^2}{N}\right] \quad (39)$$

Where $(x_i, y_i, z_i, R_i)$ are the estimated values for $i$th data set and $(x_0, y_0, z_0, R_0)$ are the original data. Then the RMS error value is found and tabulated in Table IV for both the methods. The Eberly's method show a poor fit for case 3 and case 4. This must be due to the limitation of 25 iteration steps used for convergence. Increasing the iteration steps will help for better fitting but will increase the computational cost too. We have noticed that for Eberly's method case 4 needs 250 iterations to achieve RMS error of 2.43. However for consistency of comparison for both speed and accuracy, we will keep same max iteration number for all the test cases.

Table IV: RMS error x $10^3$ for fitting

|  | Case 1 | Case 2 | Case 3 | Case 4 |
|---|---|---|---|---|
| Present work | 0.11 | 0.42 | 1.83 | 9.58 |
| Eberly [4] | 0.11 | 0.43 | 98.80 | 3789.55 |

In order to estimate the computational cost for each method, we will choose N data points and will repeat the estimation for 1000 times. N will be varied from 100 to 10000 and the time taken for estimation in an i3 processor Laptop using MATLAB is shown in figure 1.

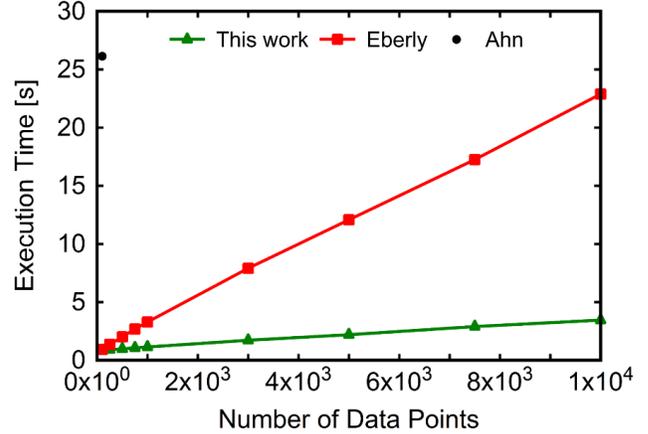

Figure 1: Speed of estimation comparison with different methods and different number of data points.

In fact the computational cost will go 10 fold higher if we increase the number of iterations to maintain the accuracy. All these estimations are done by keeping error tolerance as 0.0001 and max iterations as 25. We can see from figure 1 that the present method perform well against all other ones both in accuracy and computational time.

## IV. CONCLUSIONS

We have successfully extended the 2D version of circle fitting to 3D sphere fitting. The method shows the exact solution of fit in a geometric sense and is computationally very efficient. The algorithm was tested with multiple cases including sphere and partial spheres and shown good accuracy. The method is also compared with the existing methods to fit sphere and was shown to be robust in all sense. The MATLAB code for the present work is given in appendix 1. All other files used in this study for comparison can be obtained from author upon request.

## APPENDIX 1 - MATLAB CODE

```
function (xc, yc, zc, R) = sphere_fit(x,y,z)
(N,dum)=size(x);
Sx = sum(x);    Sy = sum(y);    Sz = sum(z);
Sxx = sum(x.*x);        Syy = sum(y.*y);
Szz = sum(z.*z);        Sxy = sum(x.*y);
Sxz = sum(x.*z);        Syz = sum(y.*z);

Sxxx = sum(x.*x.*x);    Syyy = sum(y.*y.*y);
Szzz = sum(z.*z.*z);    Sxyy = sum(x.*y.*y);
Sxzz = sum(x.*z.*z);    Sxxy = sum(x.*x.*y);
Sxxz = sum(x.*x.*z);    Syyz =sum(y.*y.*z);
Syzz = sum(y.*z.*z);

A1 = Sxx +Syy +Szz;

a = 2*Sx*Sx-2*N*Sxx;
b = 2*Sx*Sy-2*N*Sxy;
c = 2*Sx*Sz-2*N*Sxz;
d = -N*(Sxxx +Sxyy +Sxzz)+A1*Sx;
```

```
e = 2*Sx*Sy-2*N*Sxy;
f = 2*Sy*Sy-2*N*Syy;
g = 2*Sy*Sz-2*N*Syz;
h = -N*(Sxxy +Syyy +Syzz)+A1*Sy;

j = 2*Sx*Sz-2*N*Sxz;
k = 2*Sy*Sz-2*N*Syz;
l = 2*Sz*Sz-2*N*Szz;
m = -N*(Sxxz +Syyz + Szzz)+A1*Sz;

delta = a*(f*l - g*k)-e*(b*l-c*k) + j*(b*g-c*f);

xc = (d*(f*l-g*k) -h*(b*l-c*k) +m*(b*g-c*f))/delta;
yc = (a*(h*l-m*g) -e*(d*l-m*c) +j*(d*g-h*c))/delta;
zc = (a*(f*m-h*k) -e*(b*m-d*k) +j*(b*h-d*f))/delta;
R = sqrt(xc^2+yc^2+zc^2+(A1-2*(xc*Sx+yc*Sy+zc*Sz))/N);
End
```